\documentclass[conference]{IEEEtran}
\IEEEoverridecommandlockouts
\usepackage{cite}
\usepackage{amsmath,amssymb,amsfonts}
\usepackage{algorithmic}
\usepackage{graphicx}
\usepackage{textcomp}
\usepackage{xcolor}
\usepackage{cite}
\usepackage{stfloats}
\usepackage{hyperref}
\hypersetup{hidelinks,
	colorlinks=true,
	allcolors=black,
	pdfstartview=Fit,
	breaklinks=true}

\makeatletter
\newcommand{\linebreakand}{%
  \end{@IEEEauthorhalign}
  \hfill\mbox{}\par
  \mbox{}\hfill\begin{@IEEEauthorhalign}
}
\makeatother
\def\BibTeX{{\rm B\kern-.05em{\sc i\kern-.025em b}\kern-.08em
    T\kern-.1667em\lower.7ex\hbox{E}\kern-.125emX}}
\begin{document}

\title{Semi-Supervised Semantic Segmentation With Region Relevance\\
}

\author{
\IEEEauthorblockN{Rui Chen,
Tao Chen, 
Qiong Wang{*}\thanks{*Corresponding author} 
and Yazhou Yao} 
\IEEEauthorblockA{School of Computer Science and Engineering, Nanjing University of Science and Technology, Nanjing, China\\  chen\_rui@njust.edu.cn, taochen@njust.edu.cn, wangq@njust.edu.cn, yazhou.yao@njust.edu.cn} 

}
\maketitle

\begin{abstract}
Semi-supervised semantic segmentation aims to learn from a small amount of labeled data and plenty of unlabeled ones for the segmentation task. The most common approach is to generate pseudo-labels for unlabeled images to augment the training data. However, the noisy pseudo-labels will lead to cumulative classification errors and aggravate the local inconsistency in prediction. This paper proposes a Region Relevance Network (RRN) to alleviate the problem mentioned above. Specifically, we first introduce a local pseudo-label filtering module that leverages discriminator networks to assess the accuracy of the pseudo-label at the region level. A local selection loss is proposed to mitigate the negative impact of wrong pseudo-labels in consistency regularization training. In addition, we propose a dynamic region-loss correction module, which takes the merit of network diversity to further rate the reliability of pseudo-labels and correct the convergence direction of the segmentation network with a dynamic region loss. Extensive experiments are conducted on PASCAL VOC 2012 and Cityscapes datasets with varying amounts of labeled data, demonstrating that our proposed approach achieves state-of-the-art performance compared to current counterparts. Our code is available at: \href{https://github.com/NUST-Machine-Intelligence-Laboratory/TorchSemiSeg2}{https://github.com/NUST-Machine-Intelligence-Laboratory/TorchSemiSeg2}.
\end{abstract}

\begin{IEEEkeywords}
Semi-Supervised Semantic Segmentation, Pseudo-Labels, Generative Adversarial Networks
\end{IEEEkeywords}

\section{Introduction}

Image semantic segmentation is a fundamental task in computer vision, which aims to assign a semantic label to each pixel of an image. It is widely used in many fields, such as autonomous driving, medical image analysis, geographic information systems and robotics~\cite{chen2021semantically,liu2023fecanet}. Semantic segmentation is a dense pixel-level classification task that requires humans to label each pixel. Hence labeling a large-scale dataset is usually time-consuming. Many approaches have been recently proposed to alleviate the annotation burden, such as automatic dataset construction~\cite{yao2017exploiting,yao2020towards}, unsupervised domain adaptation~\cite{chen2021enhanced,pei2022hierarchical}, and weakly supervised learning~\cite{yao2021non,chen2022saliency}. In this paper, we focus on semi-supervised semantic segmentation that aims to obtain remarkable segmentation performance even with only a small amount of labeled data.

\begin{figure}[t]
            \centering
            \setlength{\abovecaptionskip}{0.cm}
		\includegraphics[width=1.0\linewidth]{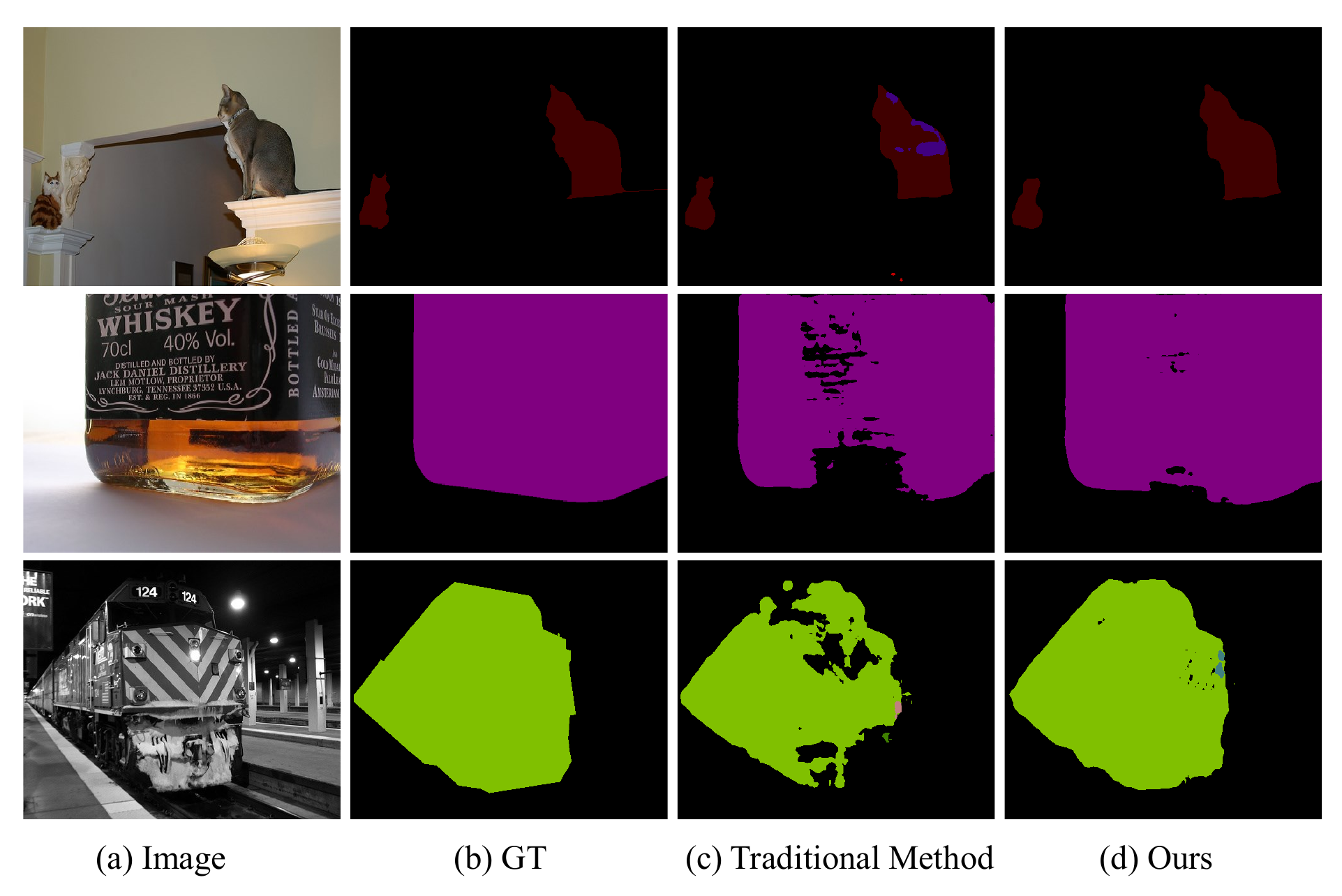}
	\caption{Comparison between traditional methods and ours. (a) Input image. (b) The Ground Truth. (c) Results of traditional methods. (d) Our results. Our proposed method can alleviate the local inconsistency in prediction and obtain more integral object segmentation.}
	\label{fig:disad}
\end{figure}

In the field of semi-supervised semantic segmentation, the dataset contains a small part of labeled data and many unlabeled images. The key point of semi-supervised semantic segmentation is exploiting the unlabeled data. Two mainstream approaches in semi-supervised semantic segmentation are consistency regularization~\cite{kim2020structured} and pseudo-labeling~\cite{lee2013pseudo}. 
Consistency regularization is based on clustering assumption and smoothing assumption. Its idea is that data with different labels are separated at low density, and similar data have similar outputs. Therefore, if the data is applied with different perturbations, they can be regarded as similar data with consistent output. Representative methods using consistency regularization are ${\pi}$ -model~\cite{laine2016temporal}, Temporal Ensembling~\cite{laine2016temporal} and Mean-Teacher~\cite{tarvainen2017mean}. The idea of pseudo-labeling~\cite{lee2013pseudo} is relatively simple: to train a model on labeled data and then use the trained model to predict the labels of unlabeled data for the pseudo-label generation. Then the unlabeled images with pseudo labels are combined with the labeled data to obtain the augmented training data. Consistency regularization and pseudo-labeling are orthogonal and CPS~\cite{chen2021semi} proposes to combine the two approaches. It employs a two-branch network to introduce different perturbations to the input, and make the output of one branch as the pseudo-label of another one for training. 
But the quality of the generated pseudo-label is usually much inferior to the ground truth. Consequently, the network will update in the wrong direction, leading to poor segmentation results.
As shown in Fig.~\ref{fig:disad}, the noisy pseudo-labels will lead to cumulative classification errors and aggravate the local inconsistency in prediction. For example, as illustrated in the first row of Fig.~\ref{fig:disad}, part of the cat region is misclassified as the dog. 

\begin{figure*}[t]
            \centering
            \setlength{\abovecaptionskip}{0.cm}
		\includegraphics[width=1.0\linewidth]{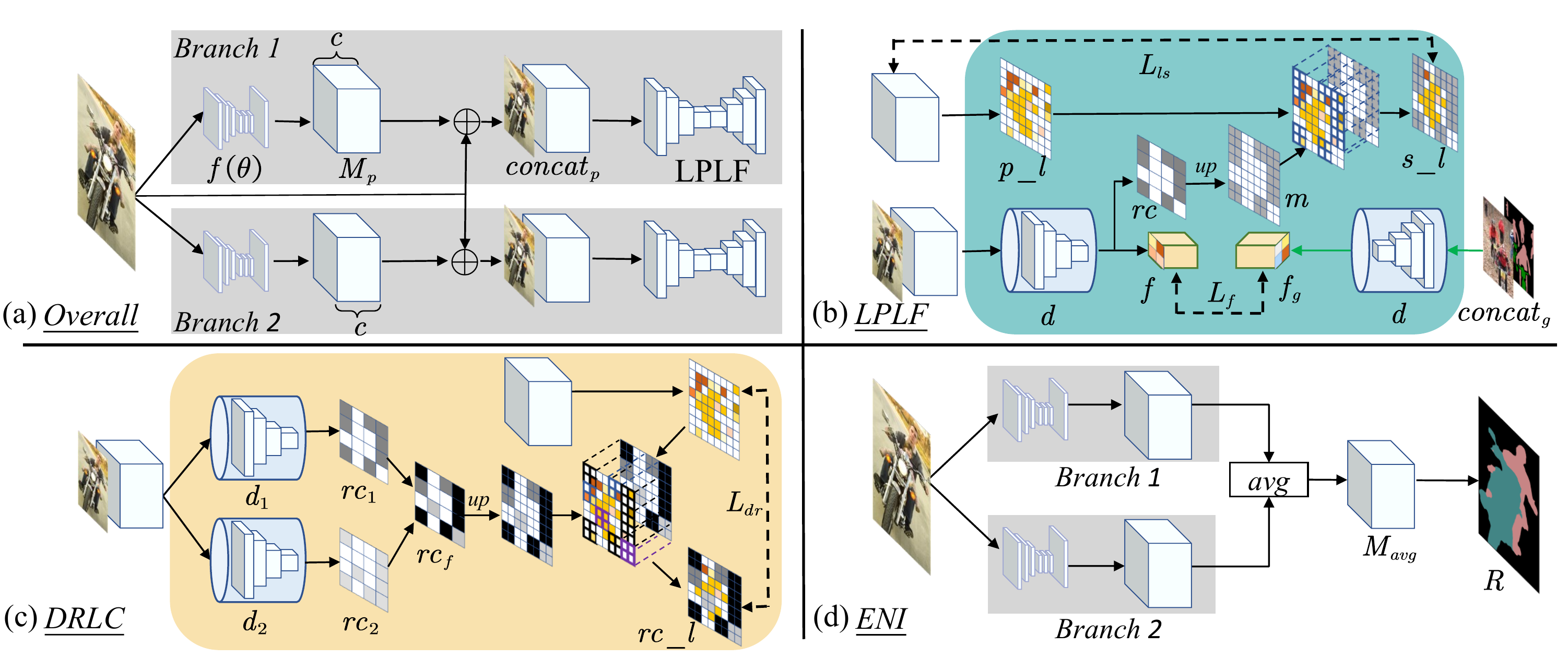}
	\caption{The architecture of our proposed approach. The "$c$" in (a) represents the c channels of $M_p$.}
	\label{fig:fig_framework}
\end{figure*}

In this paper, we propose a Region Relevance Network (RRN) to alleviate the problem mentioned above. We notice that the traditional methods only consider the prediction accuracy of a single pixel and ignore the local consistency, i.e., the relationship among pixels in a local region.
Inspired by \cite{mittal2019semi}, we propose a local pseudo-label filtering module that adopts generative adversarial networks to assess the accuracy of the pseudo-label by considering the local consistency. While the segmentation network acts as a generator, a discriminator consisting of convolutional layers is added to derive the region-consistency decision map, which indicates whether the predictions of pixels in the local region are trustworthy. We then select the reliable pseudo-labels according to the decision map for the calculation of local selection loss. In addition, to reduce pixel-level classification errors, we propose a dynamic region-loss correction (DRLC) module. It takes full advantage of the inconsistency of the two branches and further rates the reliability of the prediction map for dynamic region loss calculation according to whether the prediction map can fool the discriminator of each branch. To fully benefit from the double branches, we also fuse the prediction of the two branches
as the final segmentation map. As shown in Fig.~\ref{fig:disad}, our approach can effectively alleviate the local inconsistency problem and obtain better segmentation results.

Our contributions can be summarized as follows:
\begin{itemize}
  \item We propose a region relevance network to alleviate the local inconsistency problem caused by inaccurate pseudo-label.
  \item We propose a local pseudo-label filtering module that leverages discriminator networks to assess the accuracy of the pseudo-label at the patch level and leverage a local selection loss to mitigate the negative impact of noisy pseudo-labels.
  \item We propose a dynamic region-loss correction module, which takes the merit of network diversity to further rate the reliability of pseudo-labels. 
  \item We validate our approach on Pascal VOC 2012 and Cityscapes, and the experiments show that our method outperforms existing SOTA methods.
\end{itemize}

\section{Proposed Method}

\subsection{Overall Framework}
In this paper, we focus on the task of semi-supervised semantic segmentation that aims to learn from the labeled and unlabeled images. Our framework is illustrated in Fig.~\ref{fig:fig_framework}(a). Following the work of CPS~\cite{chen2021semi}, we forward the images into two parallel segmentation networks $f(\theta_1)$ and $f(\theta_2)$ and adopt the cross pseudo supervision strategy where the pseudo-label of one network is used to supervise the update of the other one. Considering the noise in the pseudo-label will cause the network to update in the wrong direction, we leverage a local pseudo-label filtering (LPLF) module, demonstrated in Fig.~\ref{fig:fig_framework}(b), to assess the accuracy of the pseudo-label at the patch level and select reliable labels for unlabeled images. We concatenate the prediction maps and the input to form the concatenation map ($concat_{p}$), and then we feed it into our local pseudo-label filtering module to filter low-quality pseudo-labels. LPLF module calculating local selection loss can effectively alleviate the update direction deviation problem caused by noisy pseudo-labels. While LPLF module divides regions in prediction into trusted and untrusted from a local perspective, as illustrated in Fig.~\ref{fig:fig_framework}(c), we further propose a dynamic region-loss correction (DRLC) module to adjust the credibility to more levels and adaptive weight the loss from a holistic perspective. In order to take advantage of the different learning abilities and complementariness of the two branches, we adopt the ensembling inference strategy (shown in Fig.~\ref{fig:fig_framework}(d)) to combine the predictions of both branches as the final output.

\subsection{Local Pseudo-Label Filtering}

Though the cross pseudo supervision scheme adopted in CPS~\cite{chen2021semi} can encourage the predictions across differently initialized networks for the same input image to be consistent, it does not guarantee the accuracy of the prediction. The two parallel segmentation networks will also lead to consistent wrong predictions. In other words, the noise in pseudo-labels will cause the networks to update in the wrong direction, which will aggravate the local inconsistency in the prediction of each branch. The early work of s4GAN~\cite{mittal2019semi} adopted the idea of generative adversarial networks to learn how close the prediction map is to the ground truth. However, it only generates a number to determine the reliability of the whole prediction map. Simply preserving or discarding the whole prediction will either keep much noise or undermine the potential of the pseudo-label. Therefore, we propose a local pseudo-label filtering module to consider the relationship among pixels in the local region. After assessing the accuracy of the pseudo-label at the patch level, we select reliable labels for unlabeled images. 

The architecture of our local pseudo-label filtering module is shown in Fig.~\ref{fig:fig_framework} (b). 
First, we need to map the prediction map $M_p$ in Fig.~\ref{fig:fig_framework} (a) to a prediction label $p\_l$. Then the concatenation map $concat_{p}$ is fed into the discriminator and gets two outputs: region-consistency decision map $rc$ and feature map $f$. 
Each pixel in $rc$ represents whether the corresponding region in $M_p$ is credible. We upsample $rc$ to the size of the original image to obtain a filter mask $m$.
Then according to $m$, the credible pixels in $p\_l$ are selected
to form the pseudo-label $s\_l$ for training the segmentation network. The local selection loss can be written in the following form: 
\begin{equation}
    L_{ls}=\left\{\begin{array}{l} L_{ce}\left(z, \operatorname{\S }\left(z\right)\right), \text { if } rc \geq  \varepsilon \\ 
    0, \text { if } rc<\varepsilon \end{array}\right.\label{equ:Lpl}
\end{equation}
where $z$ is the prediction and $\S$ represents the one-hot operation. $L_{ce}$ stands for cross entropy loss and $\varepsilon$ is the selection threshold.

Another output ($f$) of the discriminator is used to improve the training stability of the generative adversarial network. To alleviate the vanishing gradient problem in the original loss function of the generative adversarial network, we make the feature centers of the unlabeled image and the labeled image feature centers close in the discriminator network. As shown by the green arrow in Fig.~\ref{fig:fig_framework} (b), the concatenation map of ground truth and image ($concat_{g}$) is fed into the discriminator to generate the feature map ($f_{g}$). $f_{g}$ and $f$ are used to calculate feature matching loss:

\begin{equation}
\begin{aligned}
         L_ {f}  =  &||E_ {(x,y^*)\sim D_l} [d^{k}(concat_{g})] \\&  - E_ {u  \sim D_u}  [  [d^{k}(concat_{p})]  )]||_1
\label{equ:Lfm}
 \end{aligned}
\end{equation}
where $d^k(·)$ is the intermediate representation of the discriminator $d$ after the $k$-th layer. We use L1 loss to match the prediction features of unlabeled images and ground-truth features of labeled images.

\subsection{Dynamic Region-Loss Correction}

We have already mentioned that incorrect pseudo-labels can affect the convergence direction of the network. The labeled data with correct labels and unlabeled data with wrong labels pull the model in two directions. If we can reduce the pulling ability of noisy pseudo-labels during the training process, we can guide the model to converge to a better balance point. However, for any branch, it is not enough to measure the quality of pseudo-labels by the output of one discriminator. Therefore, we propose a dynamic region-loss correction module to improve the reliability of pseudo-label selection, which weights the loss function according to the output of the two discriminators. This module makes full use of the labeled data and the correct part of the pseudo-label for unlabeled data, and finally makes the model train toward the optimal direction.

Since the two discriminators are trained by different branches, they have different discriminating abilities. In order to fully measure whether the prediction is accurate, as illustrated in Fig.~\ref{fig:fig_framework} (c), we forward the concatenation map $concat_{p}$ into both discriminators $d_1$ and $d_2$, to get two region-consistency decision maps ($rc_{1}$ and $rc_{2}$), which indicate whether the pixels in $concat_{p}$ can fool the discriminators. 
In $rc_{1}$, the dark gray pixels represent that they are determined as wrong pseudo-labels by $d_1$. Similarly, the light gray pixels in $rc_{2}$ represent they cannot fool $d_2$. We combine $rc_{1}$ with $rc_{2}$ to generate $rc_{f}$, which divide the regions into four cases. In $rc_{f}$, the white regions represent the prediction is the most accurate and the black regions represent the most erroneous predictions. The dark gray regions and the light gray regions represent the prediction can only fool one of the discriminators.

Due to the existence of these four situations, we cannot simply use local selection loss to update the segmentation network. We propose a dynamic region loss $L_{dr}$ as follows:
\begin{equation}
      L_{dr}=
    \left\{ \begin{array}{l}

        2 * L_{ce}\left(z, \operatorname{\S }\left(z\right)\right), \text { if } rc_{1} \geq \varepsilon \text { and } rc_{2} \geq \varepsilon  \\

         L_{ce}\left(z, \operatorname{\S }\left(z\right)\right), \text { if } rc_{1} \geq  \varepsilon \text { or } rc_{2} \geq \varepsilon \\

         0,\text { if } rc_{1} < \varepsilon \text { and } rc_{2} < \varepsilon

    \end{array}\right.\label{equ:Lrl_i}
\end{equation}
where $z$ is the prediction and $\S$ represents the one-hot operation. $L_{ce}$ stands for cross entropy loss. When the prediction can fool both discriminators, its weight is set to a larger value. When the prediction can only fool one discriminator, the weight is relatively small. When the prediction cannot fool anyone, the loss is not calculated. 

\subsection{Ensembling Inference}

As the two segmentation networks $f(\theta_1)$ and $f(\theta_2)$ are with different initialization and the discriminators following them are also updated separately, the learning capabilities of $f(\theta_1)$ and $f(\theta_2)$ are quite different. If only one of the branches is used to predict, the dual-branch network cannot be fully utilized. To further exploit the different learning abilities and complementariness of the two branches, we propose an ensembling inference module to fuse the prediction of both branches as the final segmentation result during inference. We input the images into the two branches respectively, as shown in Fig.~\ref{fig:fig_framework}(d), and then perform a weighted average of the two results $M_{avg}$ to obtain the final output $R$.  

\subsection{Overall Training objective }
In the field of semi-supervised semantic segmentation, the dataset contains unlabeled images set $D_u=\{u_i\}_{i=1}^N$ and labeled images set $D_l=\{x_i,y_i^*\}_{i=1}^M$ , where in most cases $N\gg M$.
For labeled images, we compute the cross-entropy loss between the segmentation map and ground truth as the supervision signal:
\begin{equation}
	L_{sup}=\frac{1}{|D_{l}|}\sum_{x\in D_{l}}(L_{ce}(f_{1}(x),y^*)+L_{ce}(f_{2}(x),y^*))
\end{equation}
where $y^*$ is the ground truth. $f_{1}(x)$ and $f_{2}(x)$ are the prediction of $f(\theta _{1})$ and $f(\theta _{2})$, respectively. $L_{ce}$ is cross-entropy loss.

Following the work of CPS~\cite{chen2021semi}, the labeled images will also participate in the calculation of the cross pseudo-supervision loss like the unlabeled images.
\begin{equation}
	L_{cps}=L_{cps}^u+L_{cps}^l
\end{equation}
\begin{equation}
\begin{aligned}
	L_{cps}^u =\frac{1}{\left|D_u\right|} \sum_{u \in D_u} &(L_{c e}\left(f_1(u), y_{2}\right) \\
        & +L_{c e}\left(f_2(u), y_{1}\right))
 \end{aligned}
\end{equation}
where $f_{i}(u)$ is the prediction of $f(\theta _{i})$ and $y_{i}$ is the pseudo label derived from $f_{i}(u)$.

For unlabeled images, we calculate $L_{ls}$ according to Equation \ref{equ:Lpl} and calculate $L_{f}$ according to Equation \ref{equ:Lfm}.
The total loss can be expressed as:
\begin{equation}
    L=L_{s u p}+\lambda _{1} L_{cps}+\lambda _{2} L_{ls}+\lambda _{3} L_{f}
\end{equation}
where $\lambda _{1}$, $\lambda _{2}$, $\lambda _{3}$ is the weight of $L_{cps}$, $L_{ls}$ and $L_{f}$.

After introducing dynamic region-loss correction module, $L_{ls}$ is replaced by $L_{dr}$ in Equation \ref{equ:Lrl_i}, and the total loss is written as:
\begin{equation}
    L=L_{s u p}+\lambda _{1} L_{cps}+\lambda _{2} L_{dr}+\lambda _{3} L_{f}
\end{equation}
where $\lambda _{1}$, $\lambda _{2}$, $\lambda _{3}$ is the weight of $L_{cps}$, $L_{dr}$ and $L_{f}$.

\section{Experiments}

\begin{table}[t]
\centering
\caption{Comparison with previous approaches on the PASCAL VOC 2012 val set under different partition protocols. The methods marked with * are from CPS.}
\renewcommand\arraystretch{1.3}
\label{tab:voc}
\resizebox{\columnwidth}{!}{%
\begin{tabular}{lcccc}
\hline
Method& 1/16 &1/8 &1/4 & 1/2 \\            
  \hline
  MT* \cite{tarvainen2017mean}$_{\text {NeurIPS17}}$ & 66.77 & 70.78 & 73.22 & 75.41 \\
  CCT* \cite{ouali2020semi}$_{\text {CVPR20}}$
  & 65.22 & 70.87 & 73.43 & 74.75 \\
  GCT* \cite{ke2020guided}$_{\text {ECCV20}}$
   & 64.05 & 70.47 & 73.45 & 75.20 \\
  ECS \cite{mendel2020semi}$_{\text {ECCV20}}$ & -- &	70.22 &	72.60 &	74.63	\\
  DCC \cite{lai2021semi}$_{\text {CVPR21}}$
  &	70.1 &	72.4 &	74.0 &-- \\
  CPS \cite{chen2021semi}$_{\text {CVPR21}}$
 & 71.98 & 73.67 & 74.90 & 76.15 \\
  ST++ \cite{yang2022st++}$_{\text {CVPR22}}$
 &	73.2 &	\textbf{75.5} &	76.0 &	--	\\

  \hline
  Ours & \textbf{73.38} &74.91 & \textbf{76.80} & \textbf{77.04} \\
  \hline
\end{tabular}%
}
\end{table}

\begin{table}[t]
\centering
\caption{Comparison with previous approaches on the Cityscapes val set under different partition protocols. The methods marked with * are from CPS.}
\renewcommand\arraystretch{1.3}
\label{tab:city}
\resizebox{\columnwidth}{!}{%
\begin{tabular}{lcccc}
\hline
Method& 1/16 &1/8 &1/4 & 1/2 \\
  \hline
   MT* \cite{tarvainen2017mean}$_{\text {NeurIPS17}}$ & 66.14 & 72.03 & 74.47 & 77.43 \\
  CCT* \cite{ouali2020semi}$_{\text {CVPR20}}$ & 66.35 & 72.46 & 75.68 & 76.78 \\
  GCT* \cite{ke2020guided}$_{\text {ECCV20}}$ & 65.81 & 71.33 & 75.30 & 77.09 \\
  ECS \cite{mendel2020semi}$_{\text {ECCV20}}$  & -- & 67.38 & 70.70 &	72.89		\\	
  DCC \cite{lai2021semi}$_{\text {CVPR21}}$ &	-- & 69.7 &	72.7 &	--	\\
  CPS \cite{chen2021semi}$_{\text {CVPR21}}$ & 74.47 & 76.61 & 77.83 & 78.77 \\
  ST++ \cite{yang2022st++}$_{\text {CVPR22}}$ &	-- &	72.7 &	73.8 &	--		\\
  \hline
  Ours &\textbf{ 75.42} & \textbf{77.34} & \textbf{78.38} & \textbf{79.09} \\
  \hline
\end{tabular}%
}
\end{table}

\begin{figure*}[ht]
            \centering
            
            \setlength{\abovecaptionskip}{0.cm}
		\includegraphics[width=1.0\linewidth]{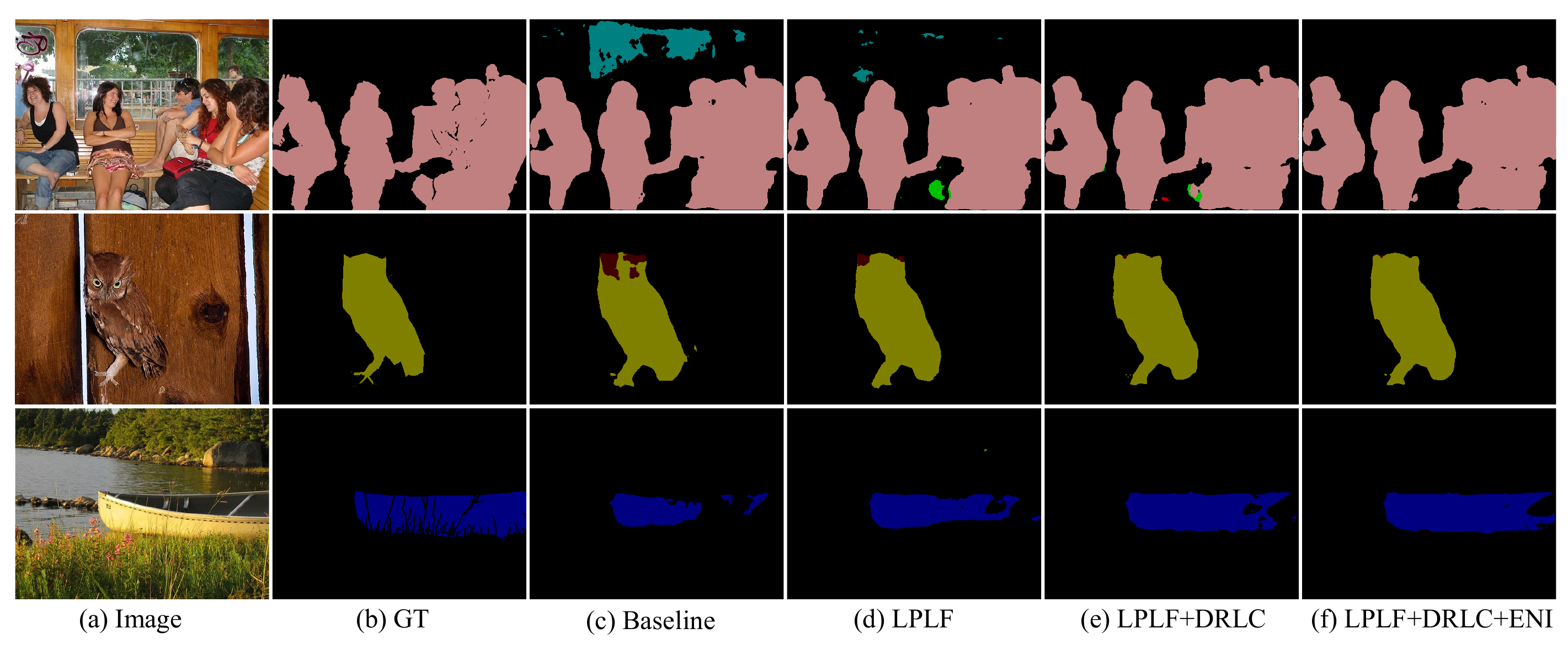}
            
	\caption{Example qualitative results from PASCAL VOC 2012. (a) Image, (b) The Ground Truth, (c) Baseline, (d) LPLF, (e) LPLF + DRLC, and (f) Our full method.}
	\label{fig_compare}
\end{figure*}

\begin{table*}[ht]
\normalsize
\begin{center}
\caption{ Element-wise component analysis under different partition protocols. LPLF: Local Pseudo-Label Filtering; DRLC: Dynamic Region-Loss Correction; ENI: Ensembling Inference.} \label{tab:ablation}
\setlength{\tabcolsep}{4.8mm}
\renewcommand\arraystretch{1.3}
\begin{tabular}{cccc|cccc}
\hline
baseline~\cite{chen2021semi} &  LPLF& DRLC& ENI &  1/16 (662) & 1/8 (1323) & 1/4 (2646)& 1/2 (5291) \\
\hline
$\surd$ & \multicolumn{1}{c}{} & \multicolumn{1}{c}{} &  & 71.98     & 73.67    & 74.90   & 76.15     \\ 
$\surd$   & \multicolumn{1}{c}{$\surd$}  & \multicolumn{1}{c}{} &    & 73.18  & 74.00 & 75.72 & 75.89  \\ 
$\surd$  & \multicolumn{1}{c}{$\surd$}  & \multicolumn{1}{c}{$\surd$}   &  & 73.03  & 74.15  & 75.81   & 76.48    \\ 
$\surd$   & \multicolumn{1}{c}{}  & \multicolumn{1}{c}{} &  $\surd$  & 73.11  & 74.00 & 75.40 & 76.30  \\ 
$\surd$  & \multicolumn{1}{c}{$\surd$}  & \multicolumn{1}{c}{$\surd$}   & $\surd$   & \textbf{73.38}  & \textbf{74.91}  & \textbf{76.80}  & \textbf{77.04}   \\

\hline
\end{tabular}
\end{center}
\end{table*}

\subsection{Experiment Setup}

\textbf {Datasets.} PASCAL VOC 2012~\cite{everingham2015pascal} is the most commonly used dataset in the field of semantic segmentation. It contains 21 target categories, of which 20 are foreground target categories, and 1 is the background category. Augmented with SBD dataset~\cite{hariharan2011semantic}, it has 10582 images for training, 1449 for validation, and 1456 for testing.  

Cityscapes~\cite{cordts2016cityscapes} contains 19 categories of targets, mainly for urban street scenes. The entire dataset consists of 50 street scenes from different cities, including 5000 accurately labeled images and 20000 roughly labeled images. We use 5000 accurately labeled images, including 2975 for training, 500 for validation, and 1525 for testing.

\textbf{Evaluation.} Following CPS~\cite{chen2021semi}, we conduct the experiment when the proportion of labeled data is 1/16, 1/8, 1/4, and 1/2. We use Mean Intersection over Union (mIoU) as our evaluation metric.

\textbf{Implementation Details.} For the two branches, Deeplabv3+~\cite{chen2018encoder} is adopted as the segmentation network. The backbone is ResNet50~\cite{he2016deep} and we use weights pre-trained on ImageNet~\cite{deng2009imagenet} for initialization. The head part of Deeplabv3+ is randomly initialized. We use mini-batch SGD as our optimizer, whose momentum is fixed to 0.9 and weight-decay is set to 5e-4. For PASCAL VOC 2012, the base learning rate is initialized to 2.5e-3, and the batch size is 6. For Cityscapes, we set the base learning rate to 0.02. Due to resource constraints, we set the batch size to 4 and update the network every two iterations. The two datasets' crop sizes are 512 × 512 and 800 × 800, respectively. The discriminator network is mainly composed of five convolution layers and one average pooling layer. It uses the ADAM optimizer~\cite{kingma2014adam}. The learning rate is initialized to 1e-4, and the beta is set to 0.9 and 0.99.
Referring to the hyperparameters in the experiment, we set $\lambda _{1}$ = 1.0, $\lambda _{2}$ = 1.0, $\lambda _{3}$ = 0.1, and $\varepsilon$ = 0.6 for Pascal VOC 2012. For Cityscapes, $\lambda _{1}$ is 5.0 and $\varepsilon$ is set to 0.7 and the other hyperparameters are the same with Pascal VOC 2012.

\subsection{Comparison with the State-of-the-art Methods}

We have conducted experiments on both Pascal VOC 2012 and Cityscapes datasets with various partition protocols and compared our proposed approach with previous state-of-the-art methods. 

\textbf{Results on Pascal VOC 2012.} Table~\ref{tab:voc} shows the results on PASCAL VOC 2012 validation set. The methods in the table are all based on Deeplabv3+. The items marked with * represent the results reproduced in CPS. The best data in each column in Table~\ref{tab:voc} is displayed in bold. It can be seen from the table that our approach performs best among all methods when the partition protocol is 1/16, 1/4, and 1/2. The mIoU values are 73.38\%, 76.8\%, and 77.04\%, respectively. Only when the partition protocol is 1/8 , the performance is inferior to that of ST++, but also 1.4\% higher than that of CPS.

\textbf{Results on Cityscapes.} In order to verify the generalization of the proposed method, we also conduct experiments on Cityscapes and the results are shown in Table~\ref{tab:city}. The results show that our method is also applicable to Cityscapes. Compared to other SOTA methods, our method achieves the best performance on Cityscapes. Especially when the proportion of labeled data is relatively small, our method is better than CPS. For example, Our method obtains 75.42\% under the 1/16 partition, which outperforms CPS by 0.95\%.

\subsection{Ablation Studies}

In this part, we demonstrate the contribution of each component proposed in our approach for semi-supervised semantic segmentation. The experimental results on the Pascal VOC 2012 dataset are given in Table~\ref{tab:ablation}. We notice that, by leveraging the local pseudo-label filtering module to assess the accuracy of the pseudo-label at the patch level for unlabeled images, we can improve the baseline performance when the partition protocol is 1/16, 1/8, and 1/4 and experience a slight performance drop at 1/2 partition protocol. Specifically, we can observe 1.2\% performance gain when 1/16 training images are labeled. We conclude that our local pseudo-label filtering module is more beneficial to situations with less labeled data where the training suffers more from the noisy pseudo-labels. When the amount of labeled data is relatively large, the pseudo-labels are more likely to fool the discriminator and may introduce noise. By introducing our dynamic region-loss correction module, we can improve the result from 75.89\% to 76.48\% when the partition protocol is 1/2. As can be seen, our local pseudo-label filtering module with a dynamic region-loss correction module can consistently improve the baseline performance. We can notice that fusing the prediction of both branches during inference further improves the segmentation performance. 

Some qualitative segmentation examples on the PASCAL VOC 2012  can be viewed in Fig.~\ref{fig_compare}. As can be seen, with our local pseudo-label filtering module to assess the accuracy of the pseudo-label and select more accurate labels for unlabeled images, we can significantly alleviate the local inconsistent prediction problem of the baseline, e.g., the head of the bird in the second row. With our dynamic region-loss correction module to further filter pseudo-labels and weight loss from a holistic perspective, we can obtain more tidy segmentation results. The adopted ensembling inference strategy can benefit from the different learning abilities and complementariness of the two branches, which further improves the segmentation.

\section{Conclusion}
In this paper, we proposed a region relevance network to alleviate the local inconsistency problem caused by inaccurate pseudo-label. 
Specifically, we first proposed a local pseudo-label filtering module, which leverages a discriminator network to assess the accuracy of the pseudo-label for unlabeled images by considering the local consistency. A local selection loss was proposed to mitigate the negative impact of wrong pseudo-labels in consistency regularization training. In addition, a dynamic region-loss correction module was proposed to take the merit of network diversity, which further rates the reliability of pseudo-labels and corrects the convergence direction of the segmentation network with a dynamic region loss. The predictions of both branches were fused during inference to obtain the final segmentation map. Experiments have shown that our proposed approach can outperform previous state-of-the-art methods.



\bibliographystyle{IEEEbib}
\bibliography{ref}


\end{document}